# Dynamic Optical Test for Bot Identification (DOT-BI): A simple check to identify bots in surveys and online processes.

By Malte Bleeker, Mauro Gotsch

## Abstract

We propose the Dynamic Optical Test for Bot Identification (DOT-BI): a quick and easy method that uses human perception of motion to differentiate between human respondents and automated systems in surveys and online processes. In DOT-BI, a 'hidden' number is displayed with the same random black-and-white pixel texture as its background. Only the difference in motion and scale between the number and the background makes the number perceptible to humans across frames, while frame-by-frame algorithmic processing yields no meaningful signal. We conducted two preliminary assessments. Firstly, state-of-the-art, video-capable, multimodal models (GPT-5-Thinking and Gemini 2.5 Pro) fail to extract the correct value, even when given explicit instructions about the mechanism. Secondly, in an online survey (n=182), 99.5% (181/182) of participants solved the task, with an average end-to-end completion time of 10.7 seconds; a supervised lab study (n=39) found no negative effects on perceived ease-of-use or completion time relative to a control. We release code to generate tests and 100+ pre-rendered variants to facilitate adoption in surveys and online processes.

**Introduction**

The reliable identification of artificially generated answers represents a mounting challenge in the collection of survey data, characterized by more sophisticated identification mechanisms and increasingly capable bots (Ferrara, 2023). Researchers have encountered challenges where, in some cases, a majority of their responses might be generated by bots (Ahler et al., 2025; Webb & Tangney, 2022). A recent study looking at bot-usage after the introduction of ChatGPT 3.0 identified an estimated 30.55% of answers recruited across varioussurveys had been generated; undermining the integrity of their results (Zhang et al., 2025). Running bots on paid survey platforms can be a lucrative business, especially when computational expenses are significantly below the expected payouts (Wang et al., 2024).

Commonly, more sophisticated identification mechanisms have also increased the burden on the participants of identifying themselves as human (Ng et al., 2025). Furthermore, some are solved by algorithmic systems, such as bots, in significantly less time than it would require humans, sometimes thereby even rendering them meaningless (Pei et al., 2020).

This paper proposes a simple bot identification mechanism that utilizes the advantage of human perception of motion. The DOT-BI consists of a "hidden" number on a background of the exact same texture that is moving and scaling differently than the number on top. The texture consists of randomly assigned black and white pixels for both the number to be identified and the background. The perception of motion makes it comparatively easy for human respondents to distinguish between the number and the background across multiple frames (i.e., a GIF or a short video). In contrast, algorithms tend to process each frame as a static image, one after the other, and are thereby unable to extract any meaning, since each single frame is only an image with randomly assigned black and white pixels. This is also underlined by two preliminary assessments—one with state-of-the-art multimodal LLMs and another with a survey of 182 participants. While none of the advanced models were able to extract the number from the GIFs or short videos, 99.5% of the human respondents in the survey were able to identify the correct number. On average, it took the respondents only 10.7 seconds to read the test instructions, view the test, identify the correct number, and submit it.

The code to generate these tests, as well as a pre-generated set of tests with 100 different values, has been made publicly available (GitHub Repository: DOT-BI, MIT License) and may be freely used and modified for private and commercial purposes.

**DOT-BI (Dynamic Optical Test for Bot Identification)**

The DOT-BI is based on the concept that an element and a background with the same texture become visually distinct only once in motion. As long as they remain in fixed positions, the element cannot be distinguished from the background. This concept, operationalized with a noise shader, is also utilized in the open-source puzzle game Motus and, combined with the need for a dataset of increased response quality, inspired the adaptation of the same concept into the area of bot identification (Jack, 2025).

Processing images is already a hurdle for bots, given that it requires more computationally intensive processing than merely interpreting text (e.g., scraped HTML) itself. Furthermore, videos require the processing of multiple images, thereby increasing the computational complexity multiple times. As a result, the limited number of available models that interpret and process videos only focus on a fraction of the total number of frames and then interpret the content for each of these frames. In the presence of elements with the same texture that only reveal themselves to the observer through different dynamic motions relative to each other, this approach, however, does not suffice. The distinct element can be text containing the message for identification (e.g., "Attention Check: Enter the number 251") or just a random value ("264"), to be entered by a user/participant. The texture consists of randomly assigned black and white colors to every pixel, to both, the background and the distinct element with the information. As a result, in a single image or frame, no content or message can be identified.

Only when the differences between the frames are assessed on a pixel level, including an adjustment for unknown scaling and directional shifts, might it become possible to detect the distinct element. While this would require a specialized algorithm for automated systems, humans can automatically distinguish between the background and the element.

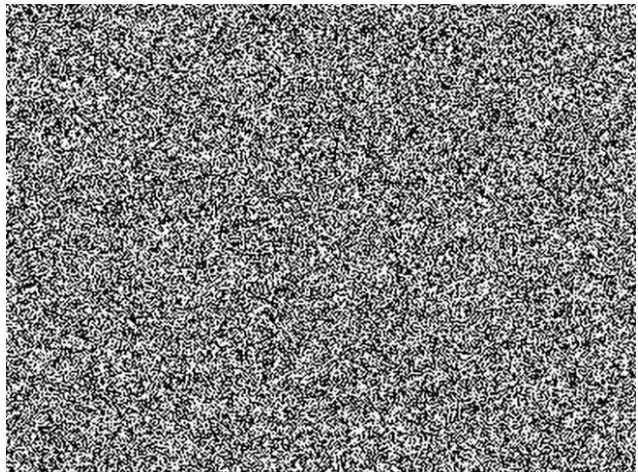

Figure 1: DOT-BI: Screenshot of a single frame with text ("243") and background having the same texture.

Only for illustrative purposes, the text element in Figure 2 is created without the same randomly applied texture. Without this, it is only through relative motion between the two elements that the message in the image can be discerned from the background by a human observer. This can be achieved by shifting the background in one direction by n pixels every frame while keeping the text in the same position. Real examples can be found in the publicly shared GitHub repository (DOT-BI, 2025).

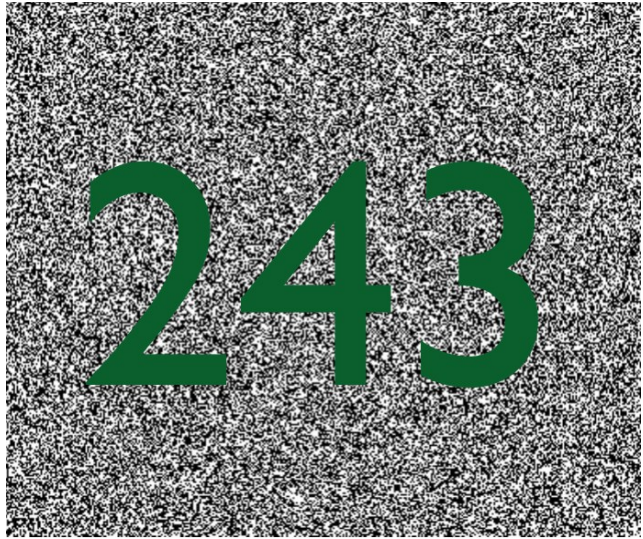

Figure 2: Text element with a different texture (green) for illustrative purposes. Notes: In the DOT-BI, the text has the same texture as the background and will only become visually distinct to the human observer in motion.

In a survey or website setting, this test can be combined with a question that would lead an automated system to guide the models toward providing a different and most likely incorrect answer with "certainty" (e.g., "Please watch the video carefully in full screen and afterward enter your favorite number").

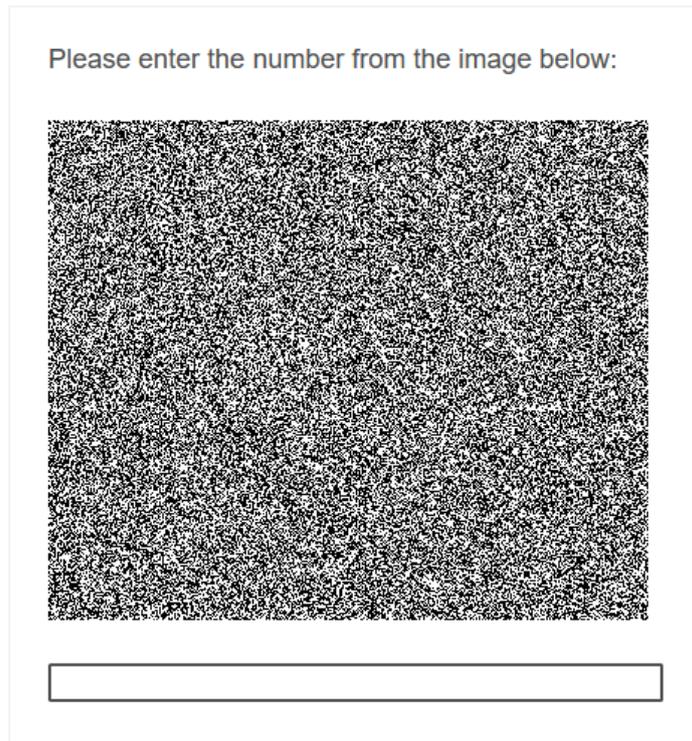

Figure 3: Example implementation of the test in a survey (GIF with the number: 243).

At the current point in time, the content/value of the message is not dynamically rendered. It can therefore make sense to include different versions (path-based or dynamically selected).

**Preliminary Assessments of the DOT-BI**

The DOT-BI was tested with state-of-the-art multi-modal models, a sample of n=182 managers (Prolific survey) and n = 39 students (observed laboratory survey). Together they aim to test three hypotheses: 1. Bots cannot identify the number shown in the video, 2. Human participants can reliably identify the number shown in the video and 3. The integration of the DOT-BI has no significant effect on completion time, reliability of answers and perceived ease-of-use of online surveys.

**Multi-Modal-Models**

Hypothesis 1 was tested using commercial licenses for the most widely available and capable models on the market. At the time of assessment, Gemini 2.5 Pro, GPT-5-Thinking were considered (some of) the most capable general models, capable of processing video input (Humanity's Last Exam, 2025). Benchmark rankings are rightfully always subject to debate; nevertheless, assessing the test with these should be highly indicative of the capability of state-of-the-art models to pass the test. Each model was tested ten times, to ensure stability in the results, with the following prompt: "In the following video is a hidden number. Identify this number." The test number is 243.

In a second approach, the underlying system was more clearly explained to the multi-modal systems. "In the following video is a hidden number. Identify this number. The background and number have the same random black and white texture but shift and scale in different directions. It is only by observing the relative change / multiple frames, that the value becomes distinguishable to the human observer."

- *Results Gemini 2.5 Pro:*
  Across all ten repetitions, this model confidently hallucinated values not included in the visuals. Following the more detailed instructions and explanation, the model fails to identify the value and keeps hallucinating. An automated system, based on this model, would have thereby failed the DOT-BI in all cases.
- *Results GPT-5-Thinking:*
  Across all ten repetitions, the model was unable to find a solution and cancelled the response. On average, responses failed after multiple minutes. During this time, a variety of code implementations were unsuccessfully tested (i.e., temporal mean, standard deviation/variance analysis, masking) to discern the value from the background. In contrast to the Gemini model, the thinking logic showed that the model itself recognized its failure to identify a value and did not hallucinate false values. Providing the explanation for the underlying mechanism did neither change the outcome nor allowed the model to identify the hidden number.

**Pilot-Assessment with an Online-Survey**

To test hypotheses 2 and 3, a field experiment was conducted first. The DOT-BI was integrated in a survey with n=182 managers (US and UK), collected via Prolific. The correct responses, as well as the time for the test completion, were collected.

- *Results Test-Survey:* 99.5% (181 of the 182) of the respondents were able to correctly solve this problem. The time from the first rendering of the instruction page and test to the respondent until the respondent's final submission of the "hidden" value was, on average, 10.7 seconds.

**Testing Ease of Use and Effect on Attention in a Supervised Setting**

A second round of surveys was conducted in a supervised, laboratory setting. 39 students from the same bachelor program were randomly assigned either to a survey with DOT-BI integration (n=27) or without (n=12) as a control group. The survey for the control group was identical – just missing the DOT-BI integration. Both surveys featured the following attention checks: (1) Asking for respondents' age in years at the beginning of the survey and their date of birth at the end of the survey, (2) an item-based attention check ("For the success of this survey, please select '7=completely agree' here) as well as (3) reverse coded items taken from an AI-acceptance survey by Ismatullaev & Kim (2024). Perceived ease of use was also measured to assess whether DOT-BI put participants into a negative frame of mind.

- *Results*: 100% of respondents in the DOT-BI group were able to correctly identify the number in the video. The average time spent on the instruction page was 21.3 seconds. However, in comparison to the control group, this amounted to no significant difference in survey completion time. A Mann–Whitney U test indicated that the difference in perceived ease-of-use and completion time between the two groups was not statistically significant at the $p < .01$ level, $U = 82.50$, $Z = 2.40$, $p = .016$, as it was high in both groups (5.8 in the control group, 6.3 in the experimental group). The failure rate in all three attention checks was low, with only one instance of attention failure across the entire study.

In summary: The DOT-BI integration had no negative effects on the perception, accuracy, and completion time of the survey with a verified human sample.

**Conclusion & Limitations**

This paper proposes a novel approach to identify bots in surveys and online processes: DOT-BI (Dynamic Optical Test for Bot Identification). The preliminary studies support this approach's difficulty to be solved by existing multi-modal models. In contrast, it was solved by 99.5% of humans in an average of 10.7 seconds. Furthermore, none of the tested state-of-the-art, video-capable, multimodal models was capable of identifying the underlying message to pass the bot identification test, even when explicitly informed of the presence of such a code within the given visuals. The code to generate one's own variants of this check, as well as one hundred pre-rendered DOT-BI variants, has been created and made available, from which researchers and developers can freely choose (Pre-generated DOT-BI Variants: GitHub Code: https://github.com/MalteBleeker/DOT-BI). The code allows for simple adjustment of parameters such as scaling and shifting directions of both the background and the element containing the visual information (i.e., number or text).

Further variations of the underlying concept with more sophisticated and dynamic relative movements may be worthwhile in the future, once bots or models have overcome this hurdle. An API that dynamically selects, distributes, and validates a respondent's choice could also be worthwhile. Extending the preliminary findings by assessing the test with different populations and other automated systems could further support or refute the suitability of the test. In addition, the risk of seizures in response to the stimuli (DOT-BI), if photosensitive epilepsy is an underlying issue in the respondent, has not yet been adequately explored. It is therefore advisable to include a warning before the test.